\documentclass{evolang9}
\usepackage{graphicx}
\usepackage{url}

\usepackage{color}

\begin{document}

\title{Why SOV might be initially preferred and then lost or recovered? A theoretical framework }

\author{Ramon Ferrer-i-Cancho}

\address{Complexity and Quantitative Linguistics Lab, TALP Research Center, Departament de Llenguatges i Sistemes Inform\`atics, Universitat Polit\`ecnica de Catalunya, Campus Nord, Edifici Omega, Jordi Girona Salgado 1-3. Barcelona, 08034, Catalonia (Spain) \\
rferrericancho@lsi.upc.edu}

\maketitle

\abstracts{Little is known about why SOV order is initially preferred and then discarded or recovered. Here we present a framework for understanding these and many related word order phenomena: the diversity of dominant orders, the existence of free words orders, the need of alternative word orders and word order reversions and cycles in evolution. Under that framework, word order is regarded as a multiconstraint satisfaction problem in which at least two constraints are in conflict: online memory minimization 
and maximum predictability. 
}

\section{Introduction}

There is converging evidence that SOV (subject-object-verb) or its semantic correlate (agent-patient-action) is a word order emerging at the very origins of language \cite{Dryer2005a,Pagel2009a,Gell-Mann2011a,Goldin-Meadow2008a}. However, the reasons why SOV is initially preferred, and later abandoned or readopted are not well understood. This requires, in our opinion, introducing a theoretical framework. 

The ordering of O, S, and V is a particular case of the ordering of a head (V) with two dependents (O and S). Let us consider a general case where a head and its $n$ dependents (complements or modifiers)\footnote{Although we are blindly borrowing the concept and head and dependent from syntactic theory, link direction or hierarchy are not relevant for our theoretical arguments.}, must be arranged linearly (the 1st elements has position 1, the 2nd element has position 2 and so on). Then, $D_l$, the online memory cost of placing the head in position $l$ ($1 \leq l \leq n+1$) is 
\begin{equation}
D_l = \sum_{d=1}^{l-1} g(d) + \sum_{d=1}^{n+1-l} g(d), 
\label{total_cost_equation}
\end{equation}
where $g(d)$ is the cognitive cost of a syntactic dependency of length $d$, which is assumed to be a strictly monotonically increasing function of $d$ \cite{Ferrer2013e}. $D_l$ is minimized when the head is placed at the center \cite{Ferrer2013e}. 
The optimal placement of the head would change if one wished to maximize the predictability of certain elements. To maximize the predictability of the head (e.g., V), the head should be placed last while to maximize the predictability of the dependents (e.g., S and O) the head should be put first \cite{Ferrer2013f}. Therefore, there is a conflict between minimum online memory and maximum predictability provided that $n > 1$. 

\section{Word order phenomena as a multiconstraint engineering problem.}

\label{discussion_section}

\subsection{The diversity of word orders}

According to the conflicts above, it is expected that thatthere is not a single
winner in the world-wide statistics of the dominant orderings for O, S and V. 
$N_1$, $N_2$ and $N_3$ are defined, respectively, as the number of languages whose dominant word order has the verb first, second or last.
In a sample of $1377$ languages, it is found $N_1=120$, $N_2=499$ and $N_3=569$ \cite{Dryer2011a}, suggesting that two possible strategies for maximizing predictability and the strategy for minimizing online memory expenditure are all three represented in world languages. Verb initial orders suggest a strategy of maximizing the predictability of the subject and the object; verb final orders suggest a strategy of maximizing the predictability of the verb; central verbs suggest a strategy of minimizing online memory expenditure.
Here it is not intended to establish whether the counts above indicate that a certain strategy is better than another in absolute terms for real languages (this is left for future work). 

Following a similar argument it has been suggested that word order diversity could emerge from the struggle between two cognitive domains (the computational system of grammar and the direct interaction between the sensory-motor and the conceptual system) trying to impose their preferred structure on human language \cite{Langus2010a}. Our approach emphasizes the conflict between universal abstract principles of sequential processing and does not need to recur to any specific cognitive system or cognitive domain.

\subsection{Languages lacking a dominant word order}

The conflict between online memory and predictability implies that there is no unquestionable placement for the verb and might explain  
the existence of a $14\%$ of languages lacking a dominant word order \cite{Dryer2011a} 
and why a lack of dominant order is an intermediate stage between SOV, i.e. maximum predictability of the verb and SVO, i.e. minimum online memory \cite{Pagel2009a}.


\subsection{Word order reversions in evolution}

The conflict between online memory and predictability may shed light on word order reversions and cycles in word order evolution.
A typical example is the reversion from SVO to SOV \cite{Pagel2009a,Gell-Mann2011a}. For instance,  Mandarin Chinese was originally an SOV language and became SVO; it is currently in the processes of moving back to SOV and thus
displays both orders \cite{Li1981a,Goldin-Meadow2008a}. Cycling between SOV and SVO could be interpreted as cycling between two incompatible attractors: maximum predictability of the head and minimum online memory. This cycling could be the manifestation of bistable system \cite{Strogatz1994}. Another example of reversion is the transition from $SVO$ to $VSO/VOS$ and then back SVO occasionally \cite{Gell-Mann2011a}.

We do not mean that reversions or cycles (repeated reversions) are a necessity of the conflict between online memory and predictability, rather, that this conflict is a relevant factor underlying word order back and forth changes. We are providing a hypothesis rather than a complete explanation. 

\subsection{Alternative orders with a head at the center}

It is well-known that SVO is an alternative word order in languages where
SVO is not the dominant word order \cite{Greenberg1963a}. This suggests that having SVO as an alternative
is a natural consequence of the conflict between maximizing
predictability and minimizing online memory: if a languages does not result in a dominant
order with the verb at the center then it should have it as
alternative to compensate the choice of a dominant order that does not
comply with all the constraints. 

\subsection{Verb last in computer prediction experiments. }

Computer simulation is a powerful tool for word order research \cite{Reali2009a,Gong2009a}.
Recently, SOV has been obtained in two-stage computer simulation experiments with recurrent neural networks (learners) that have addressed the problem of the emergence of word order from the point of view of sequential learning \cite{Reali2009a}. During the first stage, networks learned to predict the next element of number sequences and the best learners were selected. During the second stage, language was introduced and coevolved with the learners. The best language learners and the best learned language were selected. The best language learners had to comply with the additional constraint of maintaining the performance on number prediction of the first stage. 
Notice that predictability is an explicit selective pressure for the neural networks in both stages and that the languages that were selected in the second stage must have been strongly influenced by the pressure to maximize predictability. This suggests that SOV surfaces in these experiments because postponing the verb is the optimal strategy when maximizing its predictability \cite{Ferrer2013f}.

\subsection{The preference for head last in simple sequences and its loss in complex sequences.}

When a head has at least two arguments and the sender maximizes the
predictability of the last element, the last item has to be the
head \cite{Ferrer2013f}. 
Recent gestural communication experiments with only one head, i.e. a verb or an action, and two modifiers, i.e. a subject or actor and an object or patient, show a preference for placing the
head at the end in simple sequences \cite{Goldin-Meadow2008a,Langus2010a} which suggests that the predictability of the head is being maximized. Notice that the null hypothesis that there is no prior preference for head placement cannot explain this phenomenon.

A crucial finding is that this head last preference is lost in complex sequences \cite{Langus2010a}. 
The preference for head last in experiments with simple sequences and its loss in longer sequences suggests that (a) maximizing predictability dominates in short sequences (no memory-predictability conflict) while it competes with online memory for longer sequences and (b) sequence length is a critical parameter in word order phenomena. This means that predictability maximization would be the principle dominating in early stages of language evolution. 
To support hypotheses (a) and (b), it is needed to explain why online memory minimization would
have been neglected for short sequences and recovered for longer sequences. To see this, consider that 
\begin{itemize}
\item
$n=2$ is the minimum number of complements or modifiers needed to observe a preference for the verb to appear as the first or the last item of sequence. This is the number of complements or modifiers in the gestural
communication experiments where a preference of the actor (subject) and the patient (object) to precede that action (verb) has been found \cite{Goldin-Meadow2008a,Langus2010a}. 
\item
In all these experiments with short sequences, the dominant order emerging is head last (more precisely
the sequence actor-patient-action). It has been demonstrated that this placement of the
head maximizes the dependency lengths \cite{Ferrer2013f}.
\item
But as there are only $n=2$ complements/modifiers, the cost of head last is simply $D_3 = g(1) + g(2)$, the smallest among all head last configurations with $n\geq 2$ ($D_3$ is indeed the minimum value of $D_{n+1}$ for $n\geq 2$). The results of the gestural experiments suggest that the online memory cost of a head last configuration can indeed be neglected and thus only predictability matters. 
As expected, the preference for head last in gestural experiments involving longer sequences (for instance a main clause and a subordinate clause) disappears, suggesting that the conflict between online memory minimization and predictability concerns specially long or complex enough sequences
\item
However, $n=2$ does not seem to be enough to warrant that the online memory cost can be neglected since it has been shown above that about one third of word languages follow the $SVO$ or the $OVS$ order, which is a case of $n = 2$. The key is the fact that, in the gestural experiments where a preference for head last is found, elements are atomic (i.e. made of a single "word" or unit), which gives, as it will be demonstrated, a further online memory advantage with regard to the case of the ordering of subject, verb and object in word languages. So far, distance has been measured in elements (constituents) for simplicity certain elements can be made, for instance, of a subordinate clause, which happens in real languages and certain experiments \cite{Langus2010a}. Thus, online memory cost can be estimated more accurately if distance is measured in words \cite{Ferrer2008e}. In that case, the minimum online memory cost is achieved when elements are atomic (Appendix A).
Thus, the experiments in \cite{Goldin-Meadow2008a,Langus2010a} where elements are atomic, follow the setup where online memory cost can be most easily neglected, and thus, not surprisingly, head last surfaces. In contrast, subjects, objects and verbs are not necessarily atomic, which may explain why central verb orders are found in about one third of world languages (including languages lacking a dominant word order) \cite{Dryer2011a} or in the gestural experiments with complex sequences \cite{Langus2010a}, despite their {\em a priori} disadvantages in terms of predictability of the verb. Notice that the abundance of verb last orders ($N_3$) does not contradict the principle of online memory cost minimization. Indeed, the relative position of adjectives and verbal auxiliaries in verb last orders can be explained in terms of online memory cost minimization \cite{Ferrer2008e}. Thus, the fact that a language has $SOV$ as dominant does not mean that online memory cost minimization is inactive. \citeA{Langus2010a} attribute the preference for head last order in simple gestural experiments to a dissociation between communication and language but such dissociation is not necessary. Here we simply argue that a principle of language, i.e.  online memory minimization, can be neglected in those cases and thus one is able to explain the results of these experiments through another principle of language, i.e. maximum predictability. None of this principles is specific to language.  
\end{itemize}

The hypothesis of a correlation between sequence complexity and the pressure for minimizing online memory suggests the following questions: 
\begin{itemize}
\item
Why does SVO not appear more frequently in world languages? The statistical evidence indicating that SVO is the second most frequent order after SOV has already been reviewed. This raises two questions. First, do all languages have a high sequential complexity? The stability of SOV might be higher in languages producing syntactically simpler sentences. 
The issue of whether world languages have the same complexity is a matter of debate in linguistics \cite{Sampson2009a}. Second, what is the best way of measuring the abundance of a word order? It turns out that the most frequent word order, if frequency is measured in number of speakers and not in number of languages \cite{Dryer2011a}, is SVO by far \cite{Benz2010a}.
\item
Why does SVO (or the symmetric OVS) not appear more clearly in the gestural experiments in \cite{Langus2010a} with complex sentences? Evolving towards SVO from SOV may need (1) more time (2) more interaction between individuals and (3) more individuals than in the bounded experiments in \cite{Langus2010a}. We suggest that the speed of the evolution towards SVO or its accessibility may depend on at least these three factors.
The need of (1) is supported by the fact the lacking a dominant word order is a transient configuration between SOV and SVO \cite{Pagel2009a}. As for (3), notice it has been argued that the degree optimization of a language may depend on its number of speakers \cite{Sampson2009a,McWhorter2001a}.
\end{itemize}

Our interpretation of the preference for head last in simple sequences and it is loss in complex sequences given here clearly differs from that of \citeA{Langus2010a}. While  they argue that the computational system of grammar has been bypassed in simple gestural experiments showing a preference for head last, our interpretation is simply that in this case sequences are short enough to make the online memory cost negligible (Appendix). In our interpretation there is no flip of systems. The principles of sequential processing are constant but the strength of a particular principle depends on the experimental conditions.

\section{Final remarks}

It should not be interpreted that our theoretical conflict between principles predicts that all verb placements (verb first, verb last or central verb) should have about the same abundance after discarding languages lacking a dominant word order and controlling for sentence length, number of constituents, their size and other factors. The point is that a ring backbone defines the most likely transitions between orderings of $S$, $V$ and $O$ \cite{Ferrer2013e} and thus some configurations (e.g., verb initial orders) are more difficult to reach, despite their optimality, due to the initial preference for SOV and the attraction towards SVO \cite{Ferrer2013e}.

\section*{Acknowledgements}

The essence of the ideas above started circulating in January 2009 and was presented in the Kickoff Meeting "Linguistic Networks" which was held in Bielefeld University (Germany) in June 5, 2009. We thank the participants, specially G. Heyer and A. Mehler for valuable discussions. We are also grateful to E. Santacreu-Vasut for comments on the current version. This work was supported by the grant BASMATI (TIN2011-27479-C04-03) from the Spanish Ministry of Science and Innovation.

\appendix

\section*{Appendix: Online memory cost function}

$D'_l$, a more accurate online memory cost function than $D_l$, is introduced next. Imagine that there are $n+1$ constituents that can be made of more than one word and thus the sequence length in words is $m\geq n +1$. The term main head or root is used to refer to the head of the head constituent, which is the head of $V$ for the particular case of the ordering of $S$, $V$, $O$. 
Now dependencies are formed between the head word of each complement/modifier and the head of the root constituent, following the same conventions of dependency grammar \cite{Melcuk1988}. We assume that $g(d)$ is defined for $d \in [1,m-1]$. If the main head or root belongs to the $l$-th constituent of the sequence ($1\leq l \leq n+1$), the total online memory cost of the dependencies between the root and the heads of its $n$ complements/modifiers may be defined as 
\begin{equation*}
D'_l = \sum_{i=1}^{l-1} g(d_{i,l}) + \sum_{i=l+1}^{n+1} g(d_{i,l}),
\end{equation*}
where $d_{i,j}$ (with $i,j \in [1, n+1]  \subset \mathbb{N}$) is the distance in words between the head word of the $i$-th constituent and that of the $j$-th constituent. 
If constituents are made of a single word, one has $d_{i,j}=|i-j|$ and thus $D_l'=D_l$ (recall Eq. \ref{total_cost_equation}).
The point is that given a sequence of constituents, $D'_l$ is minimum when constituents are atomic (i.e. constituents are made of a single word; in that case, the only word of the constituent is the head). To see it, assume $i \neq l$ and notice that 
\begin{enumerate}
\item
$d_{i,l} \geq |i-l|$, with equality if and only if the $i$-th constituent, the $l$-th and the intermediate constituents are atomic. 
\item
$g(d_{i,l}) \geq g(|i-l|)$, since $g(d)$ is a monotonically increasing function of $d$.
\end{enumerate}

\bibliographystyle{apacite}
\bibliography{../../biblio/rferrericancho,../../biblio/complex,../../biblio/ling,../../biblio/cs,../../biblio/maths}

\begin{thebibliography}{}

\bibitem[\protect\BCAY{Bentz \BBA{} Christiansen}{Bentz \BBA{}
  Christiansen}{2010}]{Benz2010a}
Bentz, C.\BCBT{} \BBA{} Christiansen, M.~H.
\newblock{}\BBOP{}2010\BBCP{}.
\newblock{}\BBOQ{}Linguistic adaptation at work? the change of word order and
  case system from {Latin} to the {Romance} languages.\BBCQ{}
\newblock{}In A.~Smith, M.~Schouwstra, B.~de~Boer\BCBL{} \BBA{} K.~Smith\
  (\BEDS), \Bem{Proceedings of the eight international conference on the
  evolution of language}\ (\BPG\ 26-33).
\newblock{}London: World Scientific.

\bibitem[\protect\BCAY{Dryer}{Dryer}{2005}]{Dryer2005a}
Dryer, M.
\newblock{}\BBOP{}2005\BBCP{}.
\newblock{}\BBOQ{}Order of subject, object and verb.\BBCQ{}
\newblock{}In M.~Haspelmath, M.~S. Dryer, D.~Gil\BCBL{} \BBA{} B.~Comrie\
  (\BEDS), \Bem{The world atlas of language structures.}
\newblock{}Oxford: Oxford University Press.

\bibitem[\protect\BCAY{Dryer}{Dryer}{2011}]{Dryer2011a}
Dryer, M.
\newblock{}\BBOP{}2011\BBCP{}.
\newblock{}\BBOQ{}Order of subject, object and verb.\BBCQ{}
\newblock{}In M.~Dryer \BBA{} M.~Haspelmath\ (\BEDS), \Bem{The world atlas of
  language structures online.}
\newblock{}Munich: Max Planck Digital Library.
\newblock{}(Available online at http://wals.info/chapter/81. Accessed on
  2013-04-23.)

\bibitem[\protect\BCAY{{Ferrer-i-Cancho}}{{Ferrer-i-Cancho}}{2008}]{Ferrer2008%
e}
{Ferrer-i-Cancho}, R.
\newblock{}\BBOP{}2008\BBCP{}.
\newblock{}\BBOQ{}Some word order biases from limited brain resources. {A}
  mathematical approach.\BBCQ{}
\newblock{}\Bem{Advances in Complex Systems}, \Bem{11}(3), 393-414.

\bibitem[\protect\BCAY{{Ferrer-i-Cancho}}{{Ferrer-i-Cancho}}{2013\protect\BCnt%
{1}}]{Ferrer2013f}
{Ferrer-i-Cancho}, R.
\newblock{}\BBOP{}2013\protect\BCnt{1}\BBCP{}.
\newblock{}\BBOQ{}The placement of the head that maximizes predictability: an
  information theoretic approach.\BBCQ{}
\newblock{}\Bem{submitted}.

\bibitem[\protect\BCAY{{Ferrer-i-Cancho}}{{Ferrer-i-Cancho}}{2013\protect\BCnt%
{2}}]{Ferrer2013e}
{Ferrer-i-Cancho}, R.
\newblock{}\BBOP{}2013\protect\BCnt{2}\BBCP{}.
\newblock{}\BBOQ{}The placement of the head that minimizes online memory: a
  complex systems approach.\BBCQ{}
\newblock{}\Bem{http://arxiv.org/abs/1309.1939}.

\bibitem[\protect\BCAY{Gell-Mann \BBA{} Ruhlen}{Gell-Mann \BBA{}
  Ruhlen}{2011}]{Gell-Mann2011a}
Gell-Mann, M.\BCBT{} \BBA{} Ruhlen, M.
\newblock{}\BBOP{}2011\BBCP{}.
\newblock{}\BBOQ{}The origin and evolution of word order.\BBCQ{}
\newblock{}\Bem{Proceedings of the National Academy of Sciences USA},
  \Bem{108}(42), 17290-17295.

\bibitem[\protect\BCAY{Goldin-Meadow, So, \"Ozy\"urek\BCBL{} \BBA{}
  Mylander}{Goldin-Meadow \BOthers{}}{2008}]{Goldin-Meadow2008a}
Goldin-Meadow, S., So, W.~C., \"Ozy\"urek, A.\BCBL{} \BBA{} Mylander, C.
\newblock{}\BBOP{}2008\BBCP{}.
\newblock{}\BBOQ{}The natural order of events: how speakers of different
  languages represent events nonverbally.\BBCQ{}
\newblock{}\Bem{Proceedings of the National Academy of Sciences},
  \Bem{105}(27), 9163-9168.

\bibitem[\protect\BCAY{Gong, Minnet\BCBL{} \BBA{} Wang}{Gong
  \BOthers{}}{2009}]{Gong2009a}
Gong, T., Minnet, J.~W.\BCBL{} \BBA{} Wang, W. S.-Y.
\newblock{}\BBOP{}2009\BBCP{}.
\newblock{}\BBOQ{}A simulation study of word order bias.\BBCQ{}
\newblock{}\Bem{Interacion Studies}, \Bem{10}(1), 51-76.

\bibitem[\protect\BCAY{Greenberg}{Greenberg}{1963}]{Greenberg1963a}
Greenberg, J.~H.
\newblock{}\BBOP{}1963\BBCP{}.
\newblock{}\BBOQ{}Some univerals of grammar with particular reference to the
  order of meaningful elements.\BBCQ{}
\newblock{}In J.~H. Greenberg\ (\BED), \Bem{Universals of language}\ (\BPG\
  73-113).
\newblock{}London: MIT Press.

\bibitem[\protect\BCAY{Langus \BBA{} Nespor}{Langus \BBA{}
  Nespor}{2010}]{Langus2010a}
Langus, A.\BCBT{} \BBA{} Nespor, M.
\newblock{}\BBOP{}2010\BBCP{}.
\newblock{}\BBOQ{}Cognitive systems struggling for word order.\BBCQ{}
\newblock{}\Bem{Cognitive Psychology}, \Bem{60}(4), 291-318.

\bibitem[\protect\BCAY{Li \BBA{} Thompson}{Li \BBA{} Thompson}{1981}]{Li1981a}
Li, C.~N.\BCBT{} \BBA{} Thompson, S.~A.
\newblock{}\BBOP{}1981\BBCP{}.
\newblock{}\Bem{{Mandarin Chinese}: a functional reference grammar.}
\newblock{}Berkeley: University of California Press.

\bibitem[\protect\BCAY{McWhorther}{McWhorther}{2001}]{McWhorter2001a}
McWhorther, J.~H.
\newblock{}\BBOP{}2001\BBCP{}.
\newblock{}\Bem{The power of {Babel}.}
\newblock{}New York: Times Books.

\bibitem[\protect\BCAY{Mel'\v{c}uk}{Mel'\v{c}uk}{1988}]{Melcuk1988}
Mel'\v{c}uk, I.
\newblock{}\BBOP{}1988\BBCP{}.
\newblock{}\Bem{Dependency syntax: theory and practice.}
\newblock{}Albany: State of New York University Press.

\bibitem[\protect\BCAY{Pagel}{Pagel}{2009}]{Pagel2009a}
Pagel, M.
\newblock{}\BBOP{}2009\BBCP{}.
\newblock{}\BBOQ{}Human language as a culturally transmitted replicator.\BBCQ{}
\newblock{}\Bem{Nature Reviews Genetics}, \Bem{10}(6), 405-415.

\bibitem[\protect\BCAY{Reali \BBA{} Christiansen}{Reali \BBA{}
  Christiansen}{2009}]{Reali2009a}
Reali, F.\BCBT{} \BBA{} Christiansen, M.~H.
\newblock{}\BBOP{}2009\BBCP{}.
\newblock{}\BBOQ{}Sequential learning and the interaction between biological
  and linguistic adaptation in language evolution.\BBCQ{}
\newblock{}\Bem{Interaction Studies}, \Bem{10}, 5-30.

\bibitem[\protect\BCAY{Sampson}{Sampson}{2009}]{Sampson2009a}
Sampson, G.
\newblock{}\BBOP{}2009\BBCP{}.
\newblock{}\BBOQ{}A linguistic axiom challenged.\BBCQ{}
\newblock{}In G.~Sampson, D.~Gil\BCBL{} \BBA{} P.~Trudgill\ (\BEDS),
  \Bem{Language complexity as an evolving variable}\ (\BPG\ 1-18).
\newblock{}Oxford: Oxford University Press.

\bibitem[\protect\BCAY{Strogatz}{Strogatz}{1994}]{Strogatz1994}
Strogatz, S.~H.
\newblock{}\BBOP{}1994\BBCP{}.
\newblock{}\Bem{Nonlinear dynamics and chaos: With applications to physics,
  biology, chemistry and engineering.}
\newblock{}Reading, MA: Perseus Books.

\end{thebibliography}

\end{document}